\documentclass[10pt,twocolumn,letterpaper]{article}

 \usepackage[pagenumbers]{cvpr} % To force page numbers, e.g. for an arXiv version

\usepackage{microtype}
\usepackage{graphicx}
\usepackage{float}

\usepackage{booktabs} % for professional tables
\usepackage{hyperref}
\usepackage{amsmath}
\usepackage{amssymb}
\usepackage{mathtools}
\usepackage{amsthm}

\usepackage{algorithm}
\usepackage{algorithmic}

\definecolor{cvprblue}{rgb}{0.21,0.49,0.74}

\def\confName{CVPR}
\def\confYear{2025}

\title{Line of Sight: On Linear Representations in VLLMs}

\author{
\textbf{Achyuta Rajaram}\textsuperscript{1*} \quad
\textbf{Sarah Schwettmann}\textsuperscript{1,2} \quad
\textbf{Jacob Andreas}\textsuperscript{1} \quad
\textbf{Arthur Conmy}\textsuperscript{}
\\
\textsuperscript{1}MIT CSAIL\quad
\textsuperscript{2}Transluce
\\
\textsuperscript{*}Primary Author, Correspondence to achyuta@mit.edu
}

\begin{document}
\maketitle

\begin{abstract}
Language models can be equipped with multimodal capabilities by fine-tuning on embeddings of visual inputs. But how do such multimodal models represent images in their hidden activations? We explore representations of image concepts within LlaVA-Next, a popular open-source VLLM. We find a diverse set of ImageNet classes represented via linearly decodable features in the residual stream. We show that the features are causal by performing targeted edits on the model output. In order to increase the diversity of the studied linear features, we train multimodal Sparse Autoencoders (SAEs), creating a highly interpretable dictionary of text and image features. We find that although model representations across modalities are quite disjoint, they become increasingly shared in deeper layers. Project page: \url{https://github.com/multimodal-interpretability/multimodal-saes}
\end{abstract}

\begin{figure*}[!ht]
\centering
\includegraphics[width=0.9\textwidth]{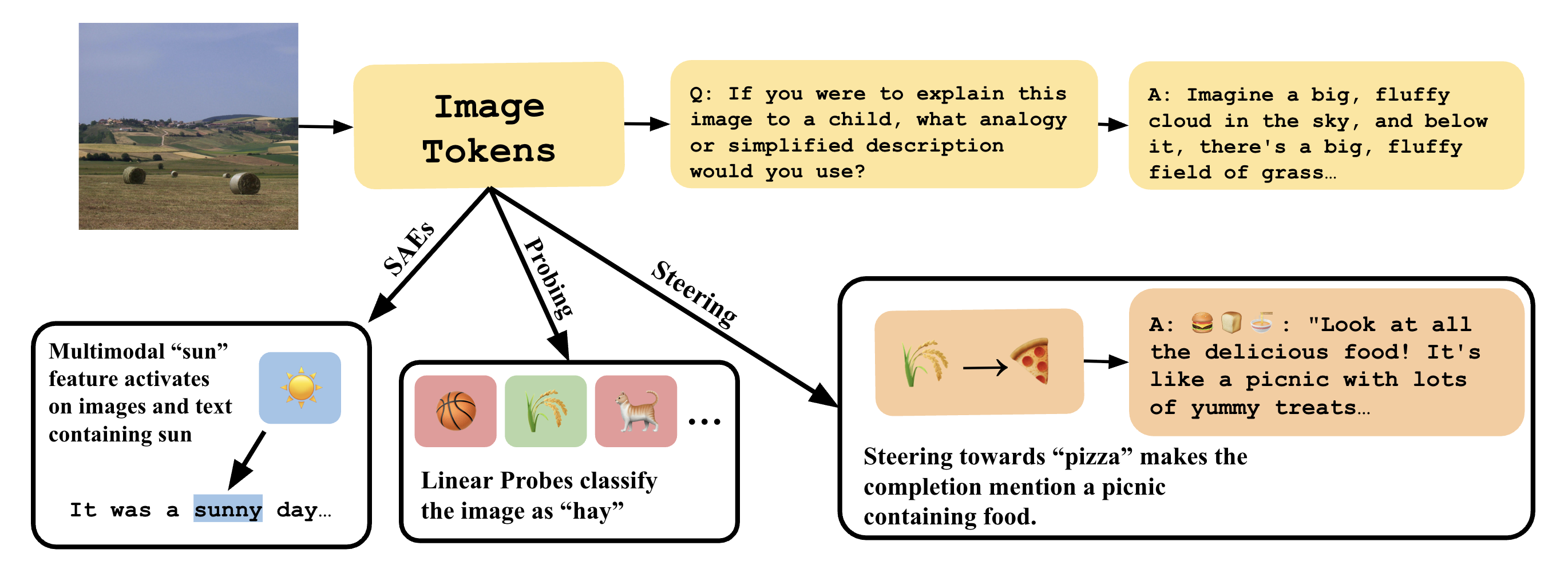}

\caption{\textbf{Linear Representations of Image Information} A schematic diagram of linear probes, steering vectors, and Sparse Autoencoders as methods for analysing linear features derived from image tokens.}
\label{teaser_figure}
\end{figure*}

\section{Introduction}

Vision-Language Models (VLLMs) that map multimodal input to text output have shown immense promise as tools in many visual tasks, from image captioning to question answering \citep{alayrac2022flamingovisuallanguagemodel, liu2024llavanext}. VLLMs are usually trained by fine-tuning a pre-trained LLM backbone on the outputs of a pre-trained image encoder end-to-end for captioning tasks. This training approach has achieved strong performance on standard benchmarks, using small amounts of multimodal data while leveraging internet-scale pre-training for image and text understanding \citep{alayrac2022flamingovisuallanguagemodel,liu2024llavanext,merullo2023linearlymappingimagetext}. 

However, LLM fine-tuning is known to be a fairly ``shallow" process, primarily enhancing existing neural circuitry \citep{prakash2024finetuningenhancesexistingmechanisms,jain2024mechanisticallyanalyzingeffectsfinetuning}. Given this, we ask: are VLLM representations of visual concepts novel and distinct from existing LLM representations, or are they simply superficial extensions?

We focus our efforts on interpreting LlaVA-Next 7b, a model trained by fine-tuning a Vicuna 7b pre-trained LLM \citep{peng2023instructiontuninggpt4} alongside a CLIP \citep{radford2021learningtransferablevisualmodels} image encoder on end-to-end multimodal instruction following, connected by a two-layer MLP projection. We extend previous work on linear representations in language-only models to study learned representations within LlaVa Next's transformer decoder, using tools including linear probes \citep{alain2018understandingintermediatelayersusing}, steering vectors \citep{turner2024steeringlanguagemodelsactivation} and Sparse Autoencoders (SAEs; \citet{andrewng,hoagy,bricken}).

First, we study how abstract image concepts are represented within the decoder-only language model, through both probing and intervention-based methods. We find that probing accurately recovers coarse image contents from the residual stream, as represented by ImageNet class, and steering allows us to edit image representations with a predictable effect on model outputs. Using causal interventions, we find that image information is transferred into text tokens in the early-mid layers of the model. Then, we study broader linear representations of text and image concepts. Using SAEs for unsupervised feature discovery, we generate monosemantic, interpretable, linear features within the model at scale. We find that linear features corresponding to ImageNet class are well-represented by the SAE basis. We also find evidence of representations becoming increasingly multimodal across layers, possibly related to the ``stages of inference" \citep{solu,stages} of the VLLM. 

\begin{figure}
    \centering
    \includegraphics[width=0.85\columnwidth]{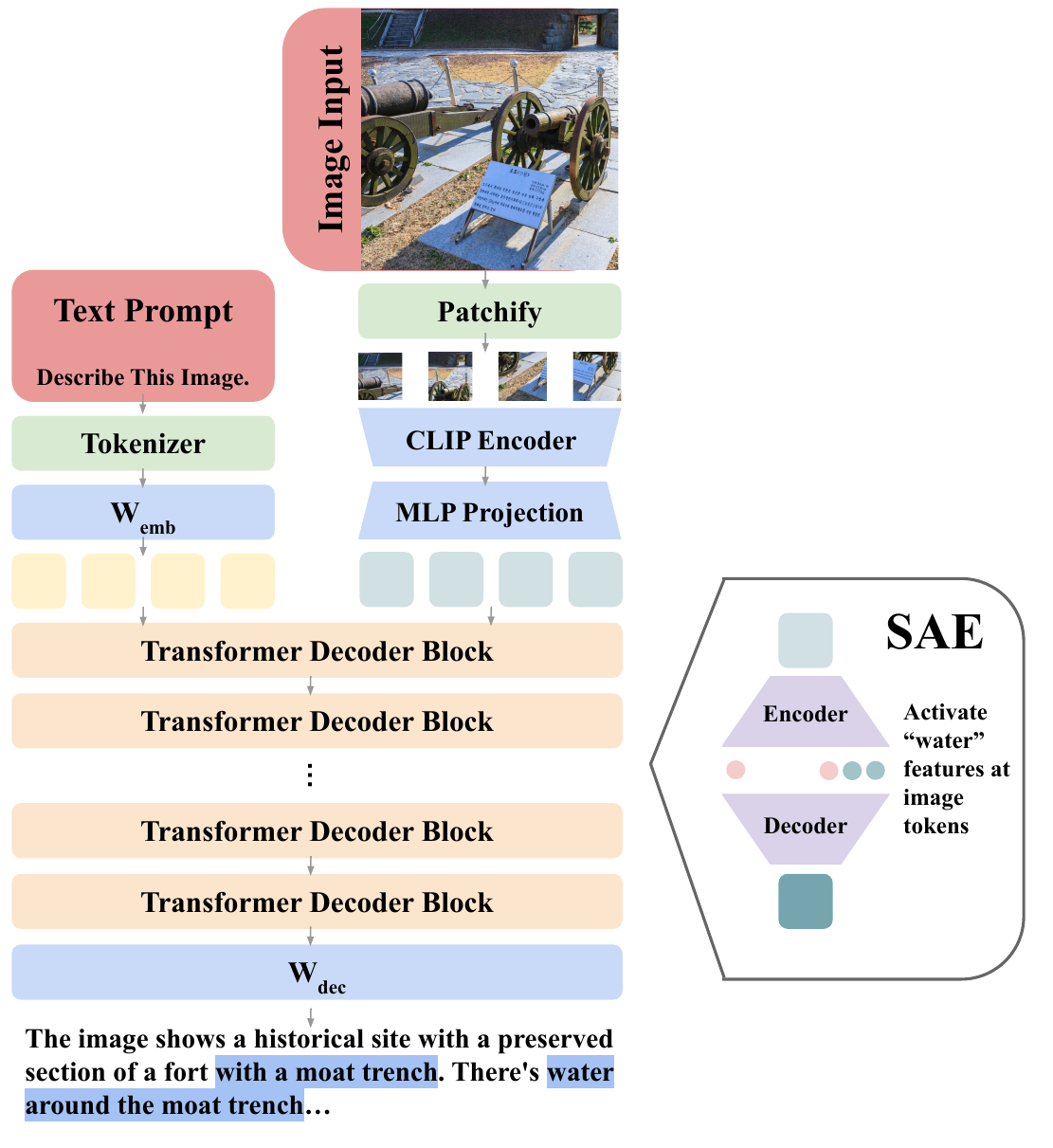}
    \caption{\textbf{Schematic of LlaVA-Next SAE Interventions:} text and image embeddings are processed separately, with the results projected into a shared latent space. If a water-related SAE feature is activated on all image tokens, the generated description of a cannon on dry land contains a hallucinated moat filled with water.}
    \label{fig:enter-label}
\end{figure}
\section{Related Work}
\subsection{Interpreting Language Models}
In recent years, extensive work has been done to interpret the internals of language models, to further understand the principles underlying model behavior. LLMs can be dissected into interpretable components in the space of weights \citep{nylund2023timeencodedweightsfinetuned}, subgraphs \citep{conmy2023automatedcircuitdiscoverymechanistic}, and activations \citep{gao2024scalingevaluatingsparseautoencoders,templeton2024scaling}. Such interpretability techniques have found use for applications such as knowledge erasure \citep{meng2023masseditingmemorytransformer}, and behavior editing \citep{wu2024reftrepresentationfinetuninglanguage}. 
\subsection{Interpreting Vision Models}

The widespread application of neural networks for computer vision has inspired efforts to understand the mechanisms behind their performance. Previous work has extensively studied the impact of individual neurons \citep{bau2020units, hernandez2022natural}, as well as found linear, distributed representations using probing \citep{alain2018understandingintermediatelayersusing}. Furthermore, such linear representations have been used to perform intervention-level analysis on generative image models \citep{xu2021linearsemanticsgenerativeadversarial}.  More recently, work has been done on finding interpretable subgraphs, as was done in language \citep{rajaram2024automaticdiscoveryvisualcircuits}, as well as the discovery of latent structure within generative models \citep{gandikota2023conceptslidersloraadaptors}.

\subsection{Interpreting Vision-Language Models}
Previous work on models like CLIP \citep{radford2021learningtransferablevisualmodels}, and LIMBER \citep{merullo2023linearlymappingimagetext} has focused on the model components which decode image information into text space \citep{schwettmann2023multimodalneuronspretrainedtextonly}. More recent work \citep{neo2024interpretingvisualinformationprocessing,jiang2024interpretingeditingvisionlanguagerepresentations} focuses on understanding VLLM representations through more sophisticated text-based methods, including the \textit{logit lens} \citep{lesswrongInterpretingGPT}, as well as localizing the locations of specific objects in token space. They find success in applications including hallucination mitigation. In this work, we focus on intervention studies instead to localize the image-to-text computation pathway, as well as explore more general linear representations of images and text within the model. 

\section{Methods}

\subsection{Linear Probes}

A hypothesis about how internal concepts are structured in decoder-only transformer language models is that they exist as one-dimensional linear subspaces within the residual stream (the output of transformer blocks) \citealp{park2024linearrepresentationhypothesisgeometry}. Such features have been used for  decoding concepts within the residual stream \citep{nanda2023emergentlinearrepresentationsworld}, and controlling model behavior \citep{turner2024steeringlanguagemodelsactivation}. 

Despite this success, several pieces of simple information appear to be encoded nonlinearly (e.g. the days of the week in \citet{engels2024languagemodelfeatureslinear}). Thus, we ask: to what extent are image semantics encoded as linear features in LlaVA-Next? To do this, we train \textit{linear probes}, linear classifiers over hidden activations, which have been used as tools for detecting the presence of encodings of concepts within models in previous work \citep{alain2018understandingintermediatelayersusing}.

We use \textit{linear probes} as classifiers for image information. Given a dataset of images $I_1, I_2, I_3, ... \in D$, where each image belongs to exactly one of $C$ classes, we compute \textit{residual stream embeddings} $R_i \in \mathbb{R}^{h_{\text{dim}}}$ by taking the mean of the residual stream activations across all image token positions at a specified layer of the language model. Using these embeddings, we use Adam \citep{kingma2017adammethodstochasticoptimization} to train a multiclass linear regression model, using a single linear layer to classify the residual stream states.

\subsection{Steering Vectors}
\label{steering_vectors}
While linear probes are useful tools for localizing information within a model, they can often find features that are not causally relevant to the model \citep{belinkov2021probingclassifierspromisesshortcomings}. If we wish to conclude that the model \textit{uses} individual vector directions to represent concepts, we need to prove that the directions have effects on model outputs. Thus, we follow \citet{turner2024steeringlanguagemodelsactivation, giulianelli2021hoodusingdiagnosticclassifiers} and use \textit{steering vectors}, a method for directly intervening on model behavior. 

As in \citet{turner2024steeringlanguagemodelsactivation}, we define \textit{positive} and \textit{negative} concepts (the class of the steering target, and the class of the source image, respectively), and sample images from each class. In keeping with our linear probing experiments, we compute \textit{residual stream embeddings} for all the images. After taking the mean of the embeddings across the positive and negative classes (random pairs of classes), we subtract the embeddings to compute the resulting steering vectors, as is done in \citet{panickssery2024steeringllama2contrastive}, a process called Contrastive Activation Addition (CAA). We then steer the model by adding in a constant vector at all image token positions at a specified layer.

In order to evaluate the resulting vectors, we attempt to steer the model to output captions relevant to the positive concept when provided an image of the negative (original) concept. A schematic of this approach is present in Figure \ref{steering_intervention}a.

\subsection{Training Sparse Autoencoders on VLLMs}

Several works have studied methods for efficient sparse codes (\citep{6287483, OLSHAUSEN19973311}); contemporary Sparse AutoEncoders (SAEs) are two-layer MLPs trained to reconstruct hidden layer activations of a larger model \citep{andrewng} in order to decompose its activations. SAEs have been used as tools for unsupervised disentanglement of computational structure in language models, from features \citep{templeton2024scaling,gao2024scalingevaluatingsparseautoencoders} to circuits \citep{marks2024sparsefeaturecircuitsdiscovering}. In this work, we seek to extend this technique to vision-language models, to allow for the discovery of image-space and multimodal features. While \citet{templeton2024scaling} find multimodal representations in Claude using text only SAEs (i.e., the golden gate bridge detector), we extend this technique to open VLLMs by training SAEs on joint text and image data, allowing us to isolate image representations.

\begin{figure}
    \centering
    \includegraphics[width=\columnwidth]{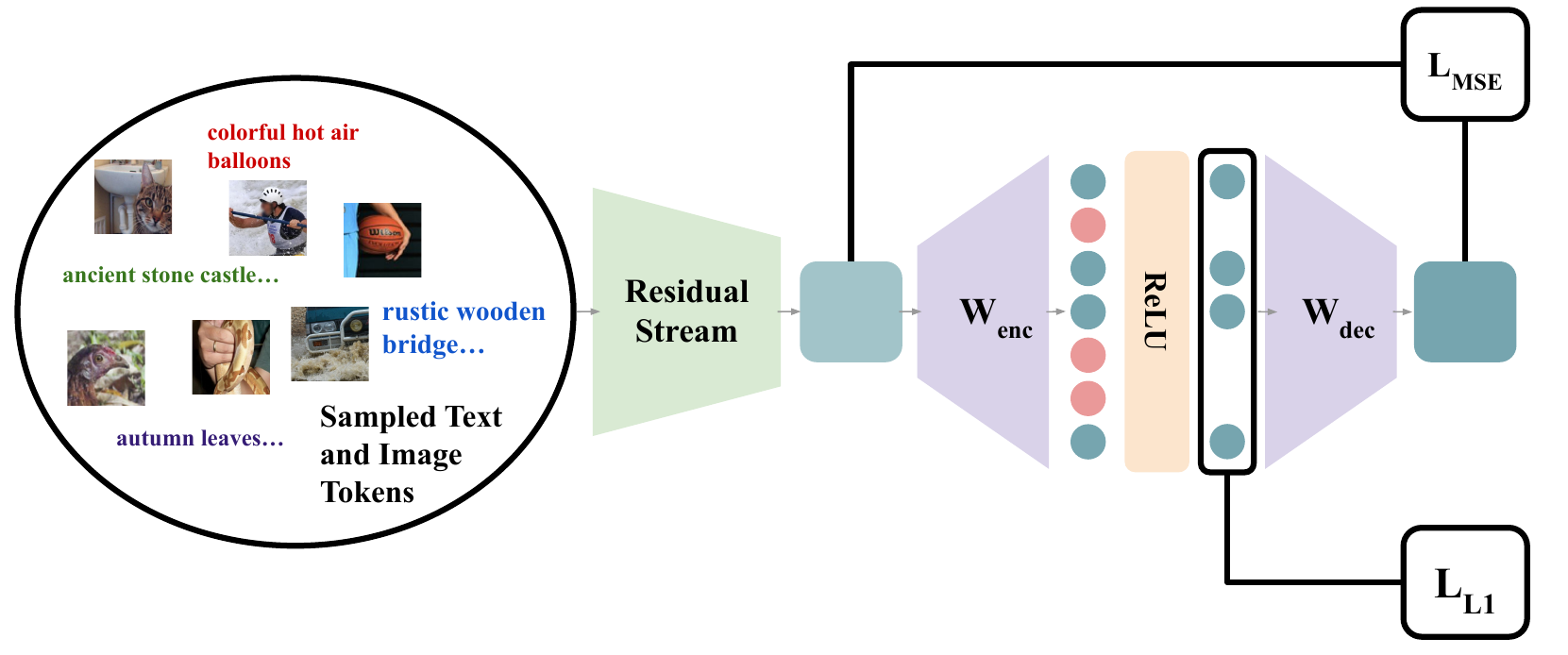}
    \caption{\textbf{Sparse Autoencoder Training:} From an image-text dataset D. we sample activations from random image and text tokens. We train a Sparse Autoencoder (SAE) to reconstruct the residual stream activations, using a sparse latent dictionary  (L1 penalized).}
    \label{fig:enter-label}
\end{figure}

Previous work trained SAEs on massive text datasets such as the Pile \citep{gao2020pile800gbdatasetdiverse}. More recently, \citet{Kissane_robertzk_Nanda_Conmy_2024} demonstrated that the distribution of learned SAE features is highly dataset-dependent. Thus to train multimodal SAEs that capture both text and image features, it is imperative to use a high-quality source of paired image-text data, as well as to balance the number of text and image tokens used. We use the training split of ShareGPT4V \citep{chen2023sharegpt4vimprovinglargemultimodal}, a diverse image-caption dataset containing 1.2 million captioned images.

We train residual-stream SAEs in the text-only decoder of LlaVA-Next, borrowing the architecture and training process from \citet{templeton2024scaling}, with no major changes. We use a ReLU activation function, and train to minimize reconstruction error (as measured by mean-squared error), as well as maximize sparsity (as approximated by the L1 across the hidden layer activations). We use \texttt{5e-5} as the learning rate with an Adam Optimizer, decaying the learning rate to zero over the last 20 percent of training. We use an 8x expansion factor, yielding 32k SAE features in the learned dictionary, and an average L0 of 5. We use a density factor, $\lambda = 5$, as is the suggested default, and linearly increase it from $\lambda = 0$ over the first 5 percent of training steps. We train for 1.5 Billion tokens at a batch size of 4096, shuffling batches for a mix of image and text tokens. We follow \citet{gao2024scalingevaluatingsparseautoencoders} and train 5 SAEs on activations from the middle layers of the model, layers 8, 12, 16, 20, 24.

\subsection{Evaluating Sparse Autoencoders}
\begin{figure}
    \centering
    \includegraphics[width=\columnwidth]{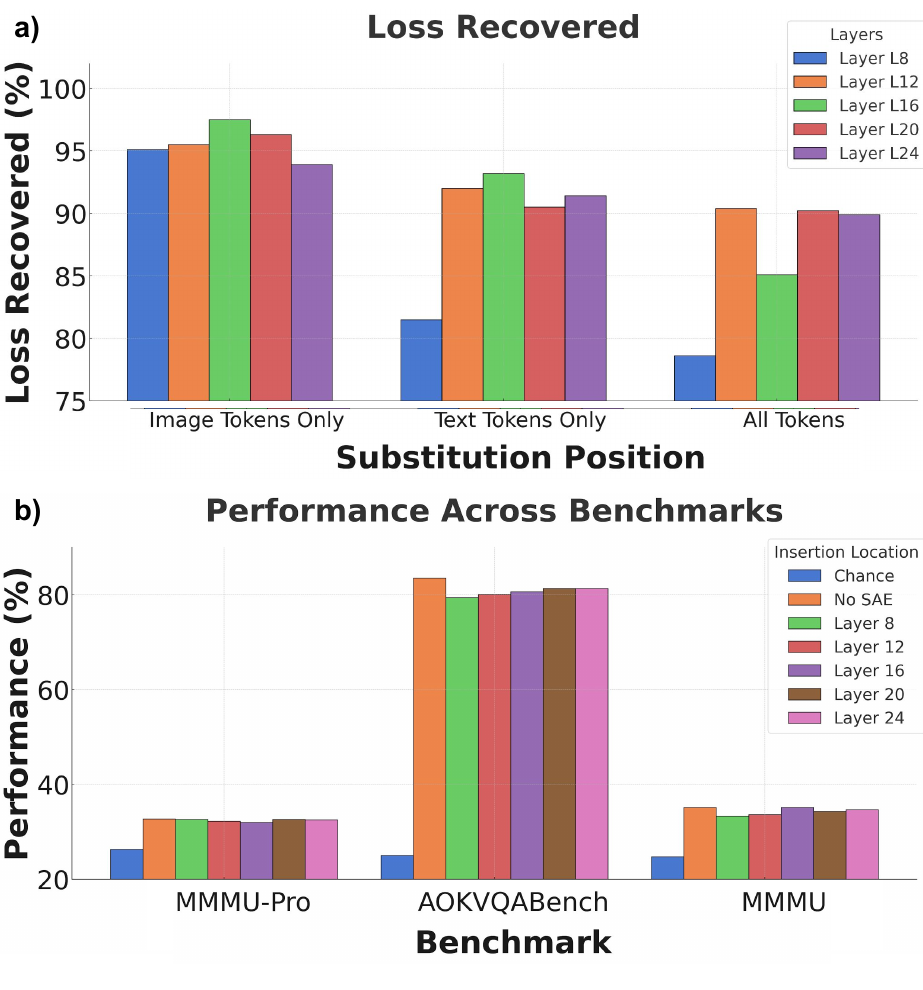}
    \caption{\textbf{SAE Evaluations}. We measure \textbf{a).} loss recovered and \textbf{b).} benchmark scores for the model with the SAE spliced in at various positions. We find that the SAEs reconstruct the information present in the image tokens well.}
    \label{sae_eval}
\end{figure}

In order to test if the trained SAEs accurately approximate model behavior, we perform evaluation of the performance and interpretability of the suite of SAEs. As was done in prior work \citep{templeton2024scaling,rajamanoharan2024improvingdictionarylearninggated}, we report loss recovered, a measure
of reconstruction fidelity which is computed by replacing activations with SAE reconstructions at a given layer and computing the fraction of CE loss recovered over a held out dataset. In our case, a held-out subset of ShareGPT-V is used, with the SAE reconstruction spliced on the text tokens, image tokens, and all tokens, at a given layer. We find that the loss recovered, compared to a baseline of zero ablation,  is fairly competitive with work on language-only models \citep{templeton2024scaling,gao2024scalingevaluatingsparseautoencoders}, (see Figure \ref{sae_eval}a).

We also evaluate the reconstruction fidelity by the use of VQA benchmarks, which provide a more comprehensive evaluation of vision-language model capabilities. Specifically, we use three popular benchmarks, MMMU, MMMU-PRO, and AOKVQA (\citep{yue2024mmmumassivemultidisciplinemultimodal, yue2024mmmuprorobustmultidisciplinemultimodal, schwenk2022aokvqabenchmarkvisualquestion}). We find that upon splicing in the SAE reconstructions for all image token features, the model only suffers a slight decrease in accuracy, illustrating that the SAEs capture visual features well (Figure \ref{sae_eval} b.). However, upon splicing in the SAE at all token positions, instead of only image tokens, the accuracy drops to near chance, with the model failing to provide a coherent response. This is likely due to a distributional shift between the SAE training data (primarily vision-language captioning, with simple instructions) and the evaluation (more complex multiple-choice question answering).

\subsection{Sparse Approximation of Steering Vectors}

While our general SAE evaluations indicate that most model behaviors are contained within the SAE, we seek to ask questions about how the SAE represents specific features. Given a concept of interest (represented by a vector in $\mathbb{R}^{h_{\text{dim}}}$), we wish the know if the SAE contains it. One way to measure the SAE's ``knowledge" of a vector, is asking: can the SAE approximate the vector well with a small set of latents? To study this, we use the SAE \textit{decoder matrix} as a sparse dictionary of vectors $\in \mathbb{R}^{h_{\text{dim}}}$. We optimize a sparse linear combination of the dictionary vectors, which reconstruct the concept of interest, as is done in \citet{Kissane_robertzk_Nanda_Conmy_2024} and \citet{rajamanoharan2024improvingdictionarylearninggated}. The sparsity of the reconstruction (measured by L0), measures the extent to which the vector lies within the dictionary.

To construct sparse approximations of a vector, we follow \citep{louizos2018learningsparseneuralnetworks,cao2021lowcomplexityprobingfindingsubnetworks}, and use the HardConcrete distribution \citep{maddison2017concretedistributioncontinuousrelaxation} in order to directly perform gradient descent, optimizing the coefficients of a linear combination on a joint objective of reconstruction accuracy (as measured by cosine similarity) and sparsity (as measured by the expected count of non-zero entries, $\mathbb{E}(||\theta||_0)$.

We can vary the sparsity of the resulting masks via an adjustable parameter $\lambda$, scaling the two terms in the loss. See Algorithm \ref{alg:HardConcreteSparse} for more details.

\begin{algorithm}[!htb]
   \caption{Sparse Dictionary Approximation via Learnable Binary Masking}
   \label{alg:HardConcreteSparse}
\begin{algorithmic}[1]
\STATE \textbf{Input:} Target vector \(v \in \mathbb{R}^{h_{\text{dim}}}\), dictionary \(W\), sparsity weight \(\lambda\), learning rate \(\alpha\), iterations \(T\), temperature \(\beta\), mask bounds \(\gamma=-0.1\), \(\zeta=1.1\).
\STATE Initialize dictionary coefficients \(x\) and mask logits \(\theta\) (both matching \(W\)'s feature dimension).
\FOR{\(t=1\) to \(T\)}
   \STATE Sample noise \(U \sim \text{Unif}(0,1)\) (same shape as \(\theta\)).
   \STATE Compute soft mask: \(S = \sigma\Bigl(\frac{1}{\beta}\bigl(\log\frac{U}{1-U}+\theta\bigr)\Bigr)\).
   \STATE Apply hard bounds: \(z = \min\bigl(1,\max(0,S(\zeta-\gamma)+\gamma)\bigr)\).
   \STATE Reconstruct: \(\hat{v} = W(x \odot z)\) where \(\odot\) denotes element-wise product.
   \STATE Compute cosine reconstruction loss: \(L_{\text{cos}} = 1-\frac{\hat{v}\cdot v}{\|\hat{v}\|\|v\|}\).
   \STATE Compute sparsity regularizer: \(R(\theta)=\frac{1}{d}\sum_{i=1}^{d}\sigma\Bigl(\theta_i - \beta\log\frac{-\gamma}{\zeta}\Bigr)\).
   \STATE Total loss: \(L = L_{\text{cos}} + \lambda\,R(\theta)\).
   \STATE Gradient updates: \(x \leftarrow x - \alpha\,\nabla_{x}L,\quad \theta \leftarrow \theta - \alpha\,\nabla_{\theta}L\).
\ENDFOR
\STATE \textbf{Output:} Sparse representation as element-wise product of coefficients \(x\) and binary mask \(z\).
\end{algorithmic}
\end{algorithm}

\section{How are Image Concepts Represented in the language model?}

One natural set of visual concepts are those defined as classes in ImageNet \citep{ImageNet}. We find that one-dimensional vectors in the residual stream represent the \textit{ImageNet class} of a given image, and show that interventions on such vectors have predictable causal effects. Furthermore, we distinguish between the features generated by probing and steering-based techniques.
\subsection{Does the residual stream contain the ImageNet class of a given image?}
\begin{figure}[ht]
\begin{center}
\vskip 0.1in
\centerline{\includegraphics[width=\columnwidth]{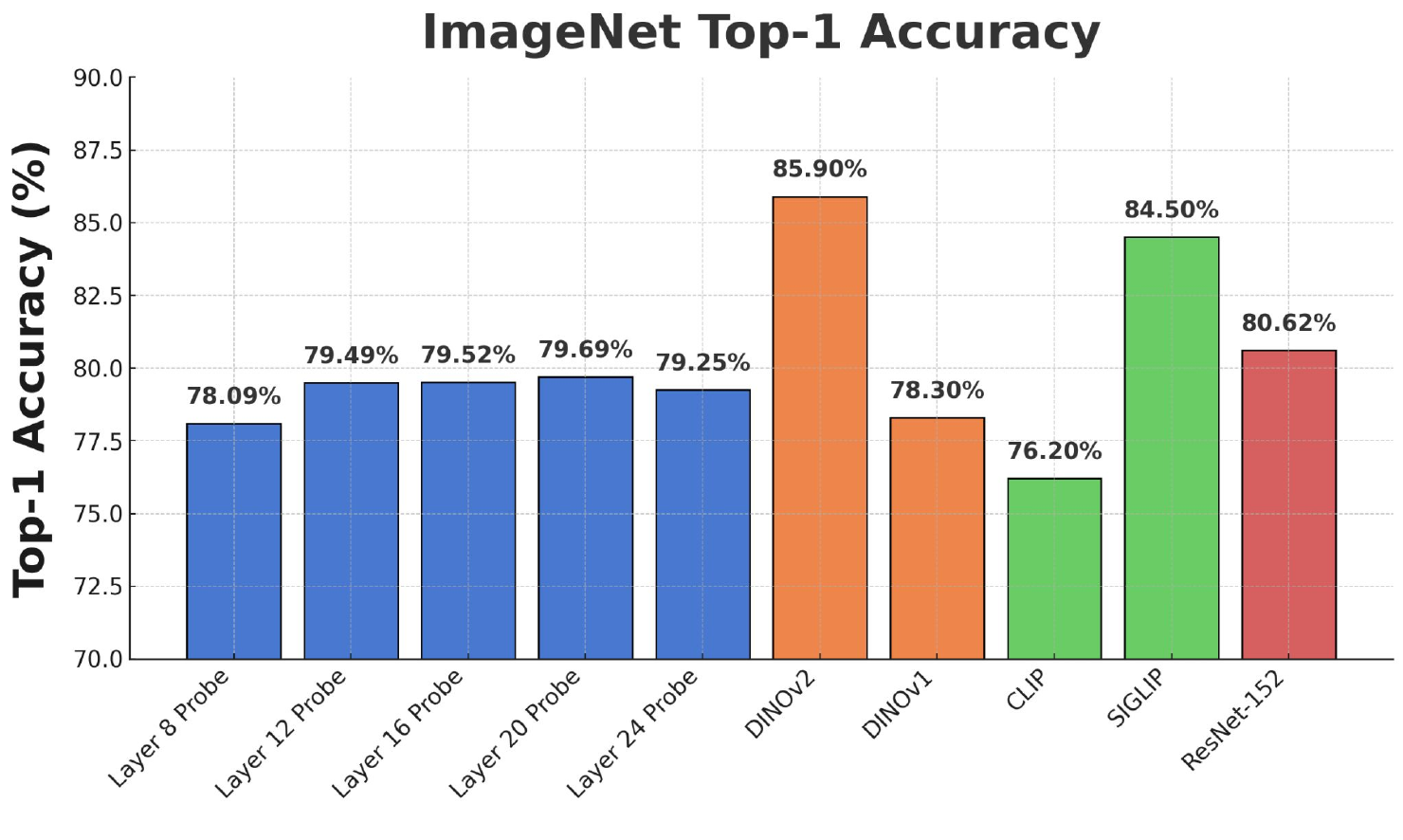}}
\caption{\textbf{ImageNet Accuracy}. We benchmark linear probes trained on mean embeddings across image tokens on ImageNet Classification. Performance is competitive with modern SSL techniques indicating linear representations of the class variable. }
\label{ImageNet Accuracy}
\vskip -0.3in
\end{center}

\end{figure}

\textbf{Yes.} We train linear probes to perform ImageNet-1K classification, using the average image token representation at a fixed layer, and compare performance of the linear probes against modern self-supervised and supervised image classifiers, namely DINO \citep{caron2021emergingpropertiesselfsupervisedvision}, DINOV2 \citep{oquab2024dinov2learningrobustvisual}, CLIP \citep{radford2021learningtransferablevisualmodels}, SIGLIP \citep{zhai2023sigmoidlosslanguageimage}, and ResNet-152 \citep{he2015deepresiduallearningimage}. We find that linear classifiers trained throughout the model achieve strong performance, on par with classifiers trained directly on self-supervised image representations (Figure \ref{ImageNet Accuracy}).
\subsection{Can the residual stream be used to steer how models see images?}
\textbf{Yes.} Previous work on steering vectors in language models \citep{turner2024steeringlanguagemodelsactivation,arditi2024refusallanguagemodelsmediated} uses a limited set of concepts, due to the difficulty of defining and collecting data for the positive and negative behaviors. As \textit{positive} and \textit{negative} concepts are easily defined in the image space, we scale our evaluation to 50 steering vectors, each defined by a randomly selected pair of ImageNet classes. We isolate steering vectors through CAA, and apply them at the layer of isolation, across several layers of the model. Some examples of the resulting model outputs are present in Appendix \ref{steering_app}.

To evaluate the efficacy of our steering vectors, we compute model completions conditioned on images of the negative class across a diverse set of questions (e.g. What are some key distinctive features or characteristics that make this thing unique?), while adding varied amounts of the steering vectors into the image tokens. As was done in \citet{chalnev2024improvingsteeringvectorstargeting}, we use gpt4o-mini to rate the model completions on \textit{coherence} and \textit{steering amount}; an optimal steering vector would steer the model outputs towards the target class while maintaining a syntactically sound output (e.g. a flawless description of an image of the target would achieve maximum score on both metrics, while an empty string would achieve minimum coherence score, but neutral steering score.) We plot a Pareto Frontier of the two scores to compare sets of vectors.

\begin{figure*}[!ht]
\centering
\includegraphics[width=0.9\textwidth]{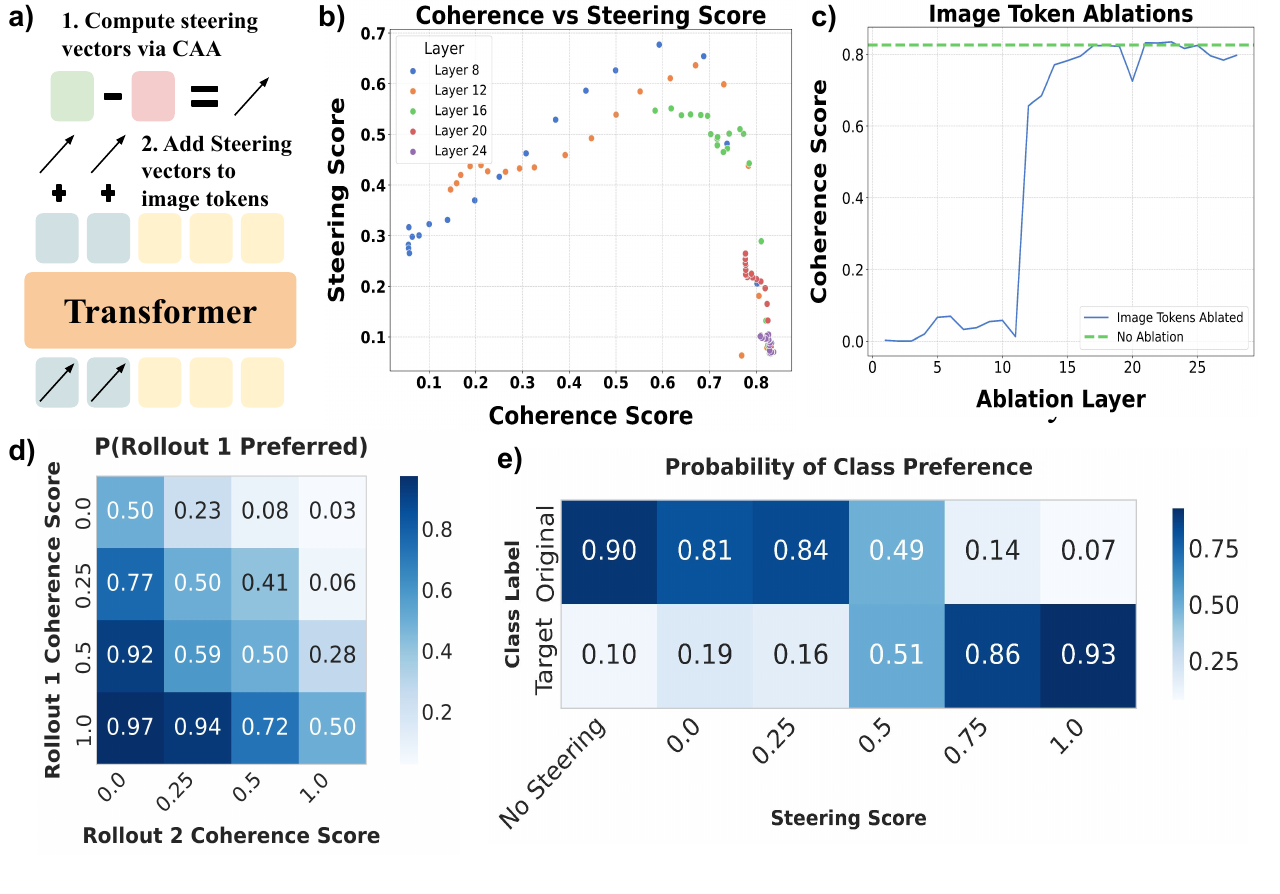}

\caption{\textbf{Steering Vector Interventions} \textbf{a).} Schematic Diagram of Steering Vectors. We compute them using activations from images of two classes, and add the resulting vectors in at image tokens. \textbf{b).} We compute coherence and steering scores as measures of steering vector efficacy. We find that steering at earlier layers is more effective. \textbf{c).} We zero-ablate all image tokens at each layer, to localize image information within the model forward pass. \textbf{d).} We find that our LLM judge's evaluations of sentence coherence align closely with human raters. \textbf{e).} We find that our LLM judge's evaluations of steering magnitude align closely with human raters.}
\label{steering_intervention}
\end{figure*}

In order to verify that our LLM Judge accurately scores model rollouts, we perform a human study of the coherence and steering scores, and measure agreement. For the coherence score, we provide pairwise model rollouts to human graders, and ask the human graders to select the more grammatically correct rollout. Plotting human preferences against LLM Judge labels, we find strong correlation (Figure \cref{steering_intervention} d). For the steering score, we provide users with a model completion as well as four potential class labels, and ask them to select the most relevant class. We again find that the human answer selections correlate strongly with the LLM judge (Figure \cref{steering_intervention} e). More details on these human evaluations are present in \cref{sv_eval}.

We perform steering across several layers, finding that targeted interventions can steer model output in a predictable fashion (Figure \ref{steering_intervention} b). As a side note, it appears that the steered outputs are not random; they are 
 highly dependent on semantic facts about the original, uncorrupted image (Figure \ref{teaser_figure}). For example, if one steers an image of hay bales on a farm towards the class ``pizza", the model describes a picnic scene, preserving that the image was taken outdoors.

 \begin{figure}[H]
\centering
\includegraphics[width=0.75\columnwidth]
{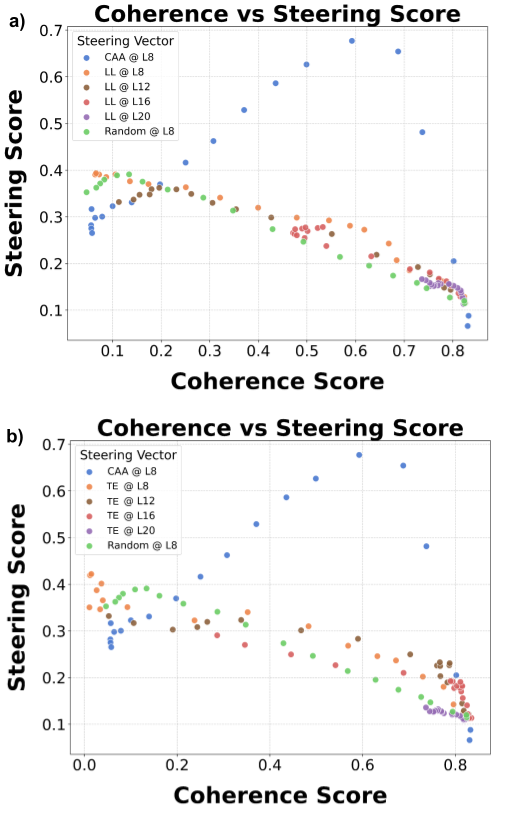}
\caption{\textbf{Text-Based Steering Vectors} We measure coherence vs. steering score for steering vectors across several steering strengths. We find that both \textbf{a).} using the unembedding matrix (Logit Lens, abbreviated LL). and \textbf{b).} using the hidden layer activations as Text Embeddings (TE) are \textit{ineffective} steering methods.}
\label{text_intervention}
\end{figure}
\subsection{Where in the model is steering most effective?}
\textbf{Early to Middle Layers.} From Figure \ref{steering_intervention} b., we see that the impact of steering vectors decreases between layers, or that interventions of the same magnitude in deeper layers have a less causal effect (as measured by our evaluations). We hypothesize that this effect is due to the model \textit{not utilizing} the image tokens at late layers of the forward pass. We verify this by performing a targeted ablation: given a layer, we zero out the residual stream at all image tokens. We then measure the impact of the ablation by computing the \textit{coherence score} of the model description(Figure \ref{steering_intervention} c). While ablating early layer image tokens is catastrophic for model performance, ablation effectiveness steeply declines halfway through the model. Ablations after layer 12 have a negligible effect on the coherence of model rollouts, indicating a well-defined location of ``information transfer".

\subsection{Can Text-derived Features Steer Image Representations?}
\textbf{Not really.} As LLaVA-Next is a vision-language model, one might wonder; does the model represent text and image features separately, or does the model use a shared code? 
\begin{figure*}[!htp]
\centering
\includegraphics[width=0.9\textwidth]{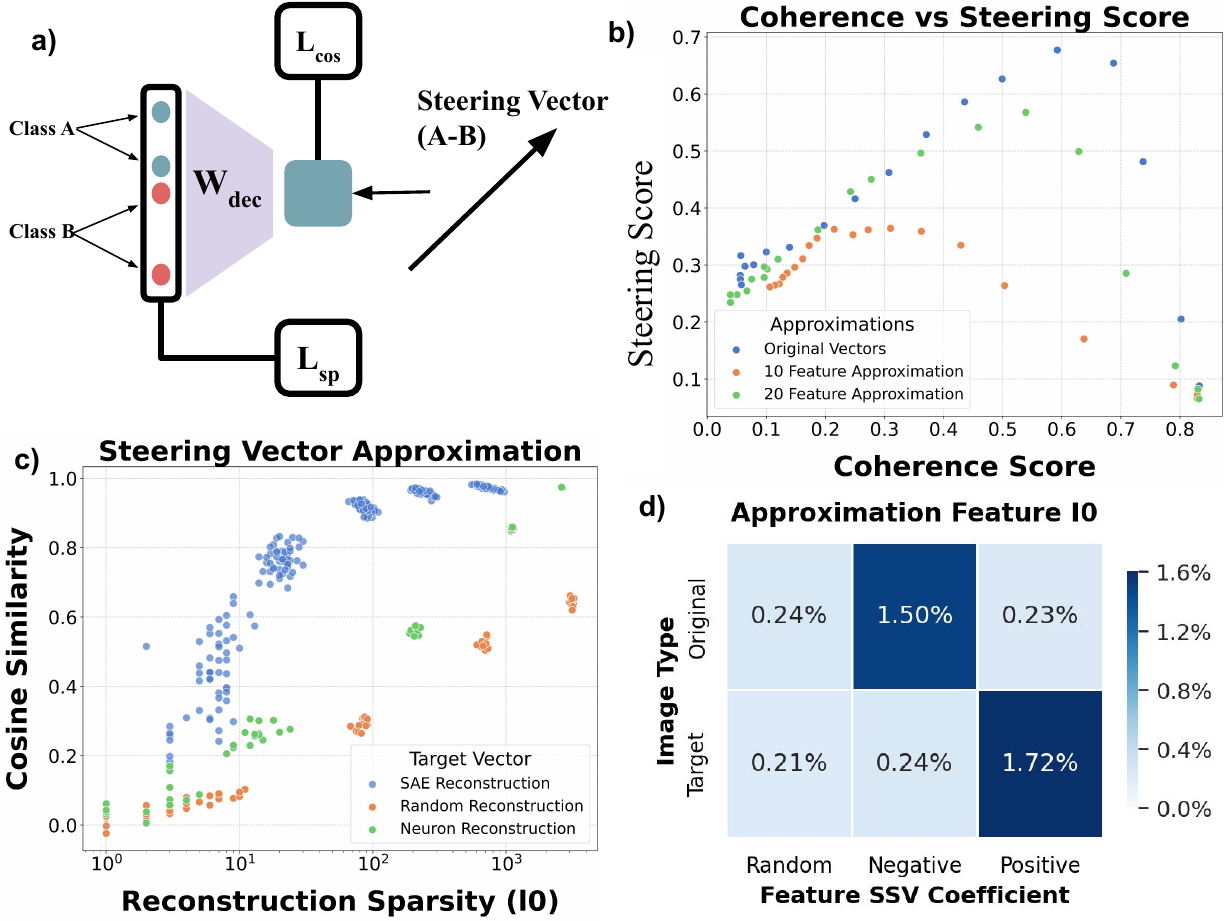}
\caption{\textbf{SAE Steering Vector Approximation} \textbf{a).} A schematic diagram of the approximation process. We optimize SAE hidden latents, using the decoder matrix as a dictionary, for sparsity ($L_{sp}$) and reconstruction accuracy ($L_{cos}$). \textbf{b).}To measure approximation quality, we compute coherence and steering score. We achieve most of the effect of the dense steering vector with only 20 SAE features. \textbf{c).} We find that SAE features can approximate steering vectors well, with neurons forming a worse dictionary. \textbf{d).} We find that the coefficients of SAE features selected by the sparse approximation process represent semantic structure in the image. }
\label{sae_approx}
\end{figure*}
Previous work has found some evidence for shared latents, with the image features being increasingly represented in the text basis throughout the model layers \citep{neo2024interpretingvisualinformationprocessing}, \citep{jiang2024interpretingeditingvisionlanguagerepresentations}. However, as we have seen in the case of linear probes, the existence of directions in a models residual stream does not necessarily imply that the model uses such directions for downstream tasks.

Thus, we perform a principled evaluation of text-based features for image token manipulation, attempting to steer model outputs with textual features. We perform two interventions, inspired by  \citet{jiang2024interpretingeditingvisionlanguagerepresentations} Logit Lens (LL), and Text Embeddings (TE). 

LL computes steering vectors for ImageNet classes A,B by finding the unembedding vectors for the corresponding text tokens, as is done in \citet{lesswrongInterpretingGPT}, subtracting them from each other, and normalizing. For TE, we instead pass the text tokens into the language model of the VLLM, extracting the residual stream at the layer of interest. We then use the resulting steering vectors to perform model interventions. We find that both of these text-derived vectors perform no better than random vectors, and are drastically outperformed by image space CAA features (Figure \ref{text_intervention}). This is unsurprising, as textual features are more present later in the model's forward pass, but the model does not use late-layer image features for captioning (Figure \ref{steering_intervention} c).

\section{What Features do SAEs recover?}
In the previous section, we explored how LlaVa-Next represents a specific type of coarse image information: the ImageNet Class. However, we are interested in answering questions about more general concepts, both within images and text. To this end, we train SAEs, finding that they uncover a diverse family of multimodal features within the model's residual stream activations. We find that even ImageNet Class is well-represented by our SAEs, pointing towards SAEs as a general tool for understanding visual representations within VLLMs.

\subsection{Monosemantic, Multimodal Features}

\begin{figure}
\centering
\includegraphics[width=0.8\columnwidth]{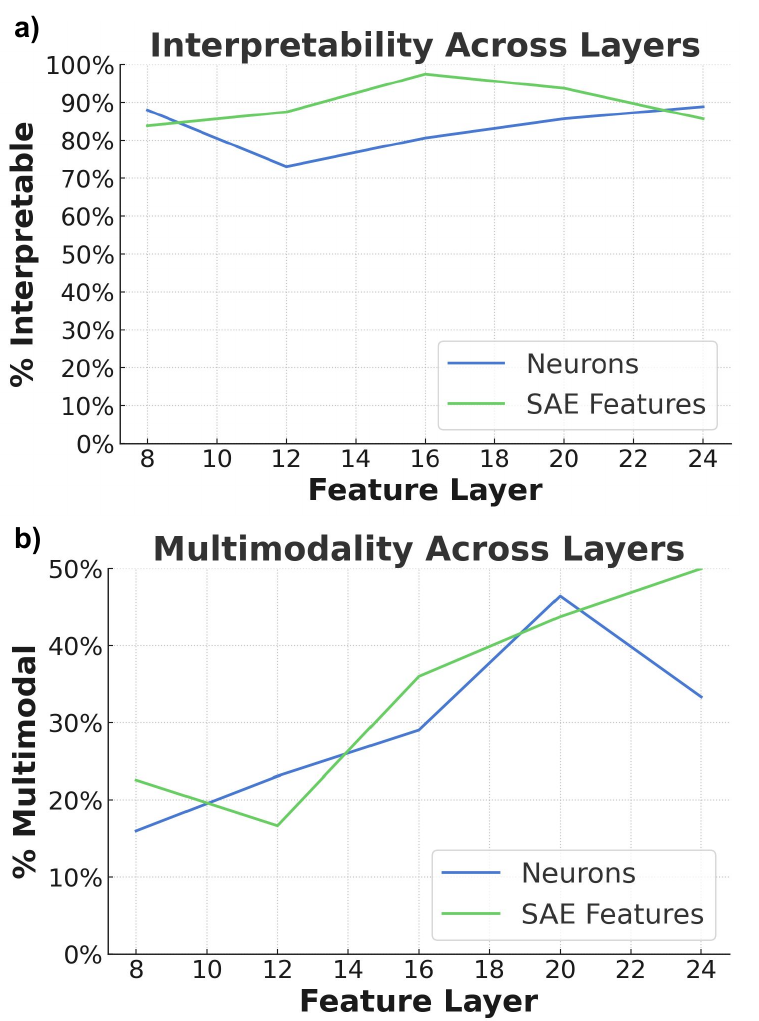}
\caption{\textbf{Manual Interpretability Experiments} \textbf{a).} We find that SAE features are slightly more interpretable than neurons, but both are highly interpretable throughout the model. \textbf{b).} We find that our human raters are more likely to label both neurons and SAE features from later layers as ``multimodal."}
\label{manual_interp}
\end{figure}

Previous work on SAEs in LLMs have found their recovered features to be highly interpretable and monosemantic, leading to their use as a tool for disentangling residual stream activations \citep{templeton2024scaling}. To this end, we benchmark the interpretability of our multimodal SAEs across image and text activations, using manual graders as is done in \citet{rajamanoharan2024improvingdictionarylearninggated}. We present a group of expert human raters with the activating examples of a randomly selected feature. The rater then decides if the feature has monosemantic text and image activations. If the human labeler finds that the descriptions match, a feature is deemed ``multimodal". We do the same process for all neurons. We find that SAEs are slightly more interpretable than neurons (\cref{manual_interp} a), and that the proportion of multimodal features (and neurons) increases throughout the model (\cref{manual_interp} b). Furthermore, we find that the neurons contain far more text-only features than the SAEs. More details on this evaluation are present in section \ref{mm_eval}.

\subsection{ImageNet Class Features}

We seek to measure how well the internal variable of ImageNet class is represented by the SAE. To approximate the internal representation of ImageNet class, we use the steering vectors outlined in Section \ref{steering_vectors}, as we have shown that they form faithful approximations. We then perform the SDA algorithm as outlined in Algorithm \ref{alg:HardConcreteSparse}, measuring the resulting sparsity. We find that the steering vectors are much ``easier" to approximate than random vectors, requiring less than 100 SAE features to approximate well (Figure \ref{sae_approx} c). Furthermore, we evaluate the resulting reconstructions on steering, as was done in section 2, and find that approximations with as few as 20 features perform comparably against the original steering vectors (Figure \ref{sae_approx} b).

Finally, it is natural to ask: are the resulting sparse approximations composed of interpretable features? To answer this, we first group the features ``selected" by the reconstruction (those with nonzero coefficient) into those positive and negative coefficients. When we consider their activation patterns across images, we find that these positive and negative features correspond directly to the positive and negative classes of the approximated steering vector, with predictable activations on images of the corresponding class (Figure \ref{sae_approx} d). 

\section{Discussion}
Through probing, steering, and SAE training, we find that diverse image-related variables exhibit linear structure; we use this to achieve fine-grained control of VLLM behavior. We also find evidence for multimodal representations organized in an ``information hierarchy" across layers. Unlike in language models, we find that neurons form a highly interpretable basis over image information, comparing favorably to SAE features. We also find that linear representations have limitations—they provide an unnatural basis for image-space adversarial attacks (Appendix \ref{Adversarial Attacks}). Future work should investigate the mechanisms underlying image-to-multimodal representation conversion and evaluate these findings across diverse VLLM architectures, particularly cross-attention-based models.
\section{Acknowledgements}
We are grateful for the support of the MIT-IBM Watson
AI Lab, and ARL grant W911NF-18-2-0218. We thank David Bau, Caden Juang, and Yolanda Xie for their useful input and insightful discussions.

\clearpage

{
    \small
    \bibliographystyle{ieeenat_fullname}
    \bibliography{main}
}

% WARNING: do not forget to delete the supplementary pages from your submission 
% \input{sec/X_suppl}

\clearpage
\appendix
\onecolumn
\section{Steering Vector Evaluation Details}
\label{sv_eval}
We evaluate steering vectors on steering model answers to general VQA questions. To do this, we select 200 random pairs of ImageNet Classes, and steer from the first element of each pair to the other. We ask the model to answer the following questions with a 200-token completion, for each input image:
\\
\begin{enumerate}
  \item [INST] what is shown in this image? [/INST]
  \item [INST] is this image related in any way to the concept \{cls[cls\_index]\}? [/INST]
  \item [INST] is this image related in any way to the concept \{cls2[cls\_index]\}? [/INST]
  \item [INST] Describe the primary purpose or function of the entity shown in this image? [/INST]
  \item [INST] If you were to explain this image to a child, what analogy or simplified description would you use? [/INST]
  \item [INST] In what environment or setting would you most likely encounter this? [/INST]
  \item [INST] what are some key distinctive features or characteristics that make this thing unique? [/INST]
  \item [INST] How large is the main subject of this image typically? [/INST]
  \item [INST] Is this image more related to the concept \{cls1[cls\_index]\} or the concept \{cls2[cls\_index]\}? [/INST]
\end{enumerate}

In order to evaluate the efficacy of our steering vectors, we use a gpt4o-mini LLM judge with the following prompts, for measuring steering amount and textual coherence respectively:
\begin{figure*}[H]

\begin{center}
\centerline{\includegraphics[width=\textwidth]{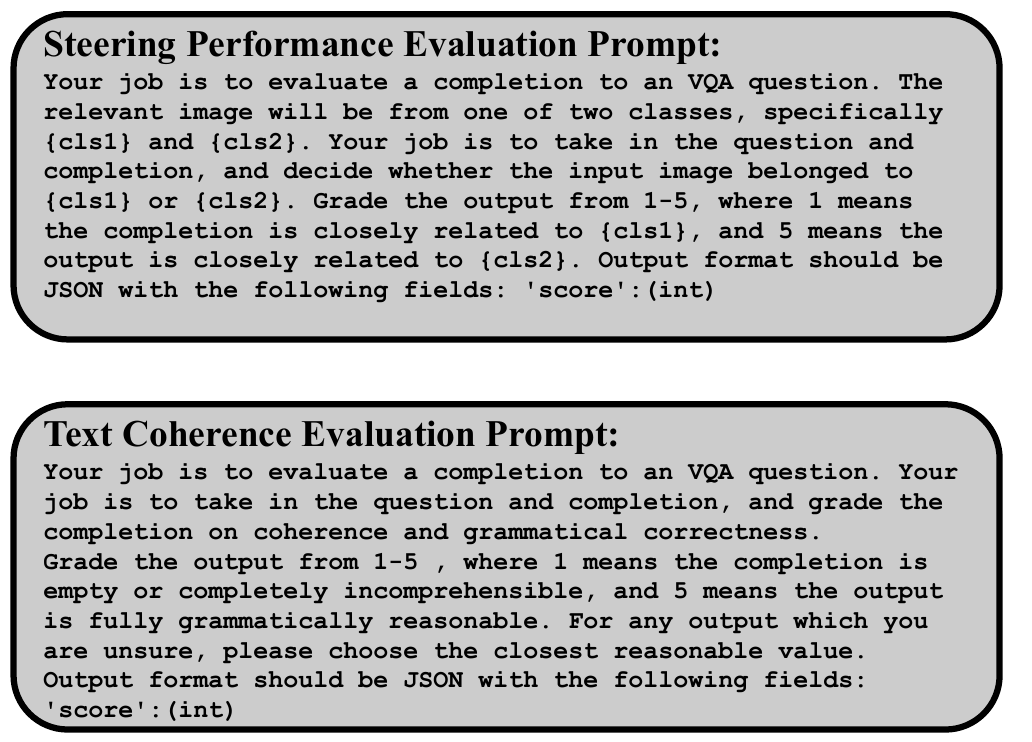}}
\label{text_intervention}
\end{center}

\end{figure*}
We normalize scores from the autograder to the range 0-1 with min-max scaling, to use in our evaluation. 

In order to verify that this autograder works properly, we perform a human evaluation using Amazon Mechanical Turk. We perform two human evaluations, one for the coherence score, and one for the steering performance score. 

\textbf{Coherence Score:} We sample 100 examples from each coherence grade (from the computed rollouts from the steering experiment), for a total of 600 samples (100 controls, with no steering applied). We then compute a dataset of paired samples, using answers to the same question next to each other. We ask our mechanical turkers to select the most coherent response, getting 10 labels per prompt. Below is an example question asked to the graders.

\begin{figure}[H]
\begin{center}
\centerline{\includegraphics[width=\textwidth]{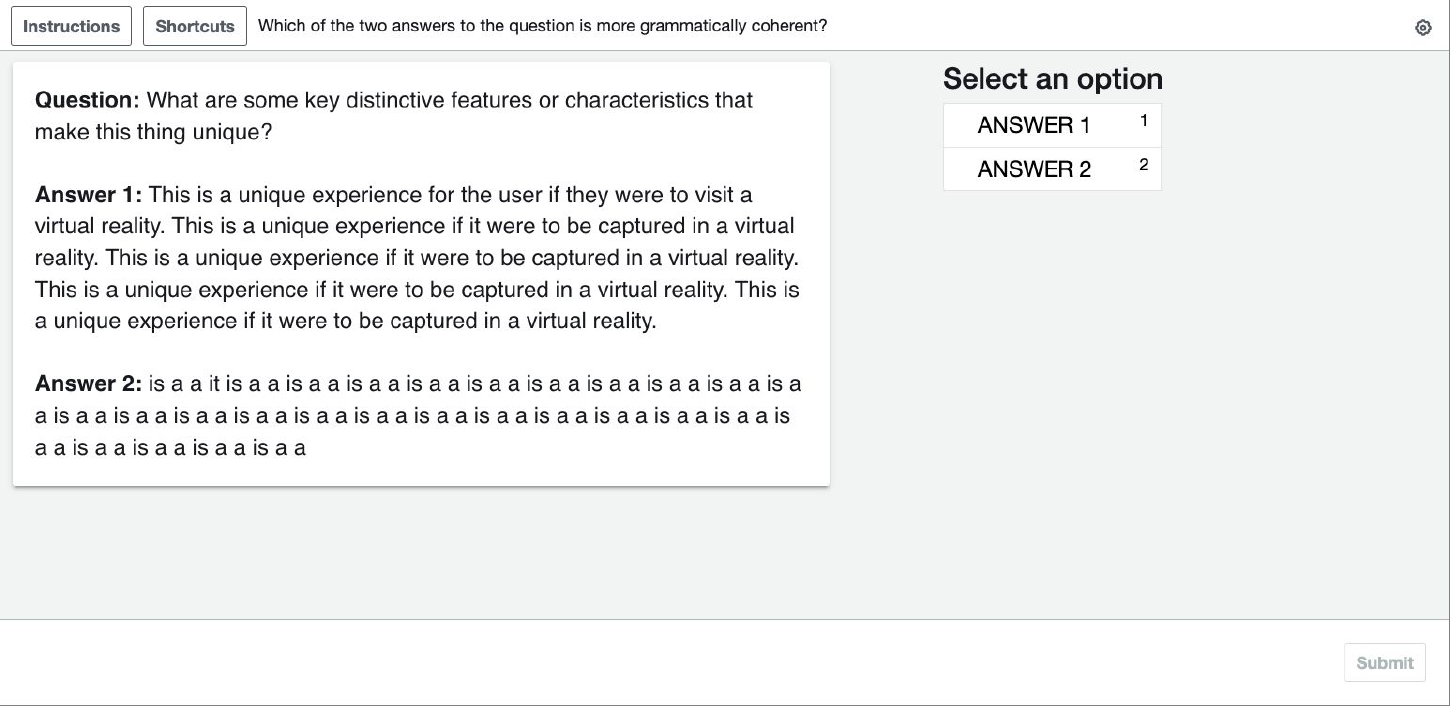}}
\caption{\textbf{Example of Coherence Evaluation} A random sample from our dataset. Clearly, Answer 1 is more coherent.}
\label{text_intervention} 
\end{center}

\end{figure}
Then, we compute the probability that the turker will prefer one answer to another, based on their respective grader scores.
\newline
\newline
\textbf{Steering Score:} To evaluate this, we first sample rollouts at each level of steering, as measured by our LLM grader. Then, we present the question-answer pairs to the human evaluators, asking them to select what concept aligns most closely with the given response from a list of four classes (the target class for the steering vector, the original class of the image, and two distractors). An example of this task is given below.
\begin{figure}[H]
\vskip 0.1in
\begin{center}
\centerline{\includegraphics[width=\textwidth]{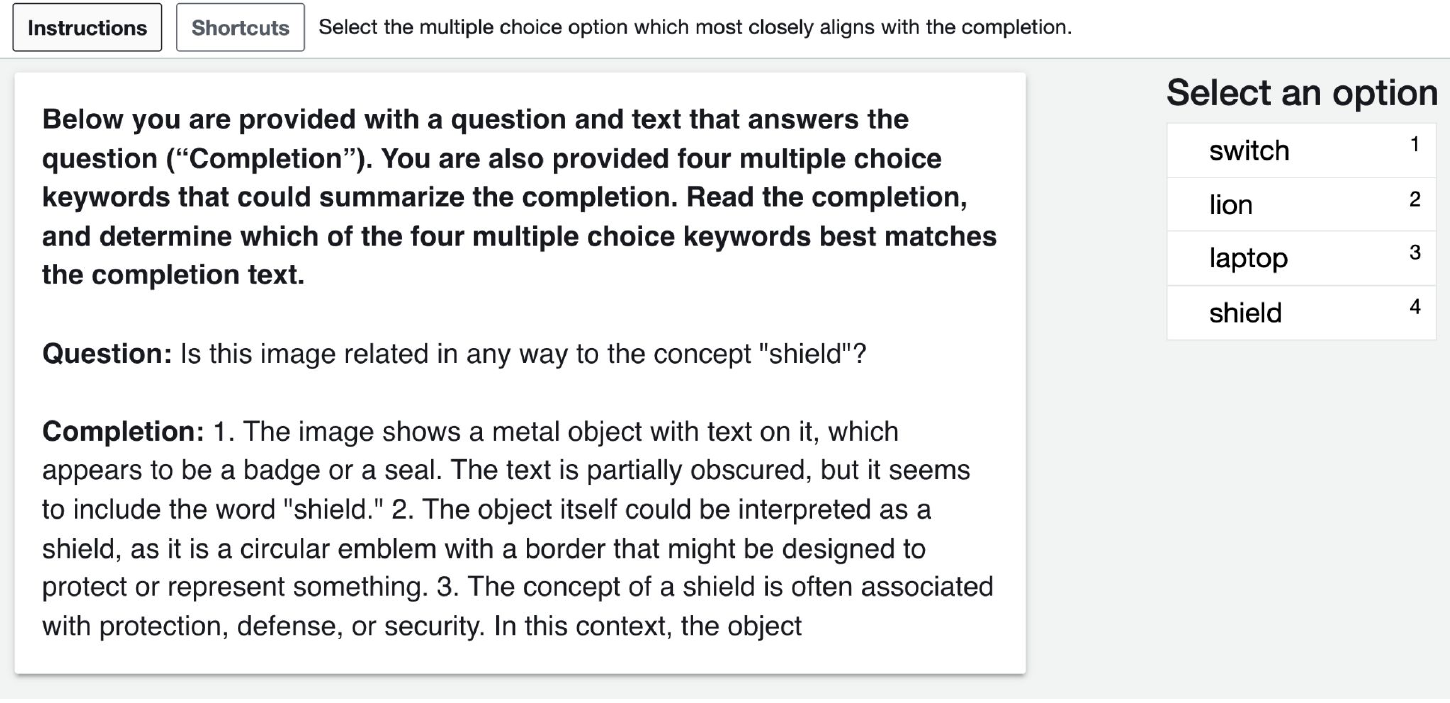}}
\caption{\textbf{Example of Coherence Evaluation} A random sample from our dataset. Clearly, the completion is most aligned with the class ``shield"}
\label{text_intervention}
\end{center}

\vskip -0.1in
\end{figure}

We then compute the probability that the user selects the target and original class, given a steering score for the prompt. Using this, we find that our LLM autograder aligns well with human data.

\section{Examples of Steering at Various Strengths}
\label{steering_app}
\begin{figure}[!htp]

\begin{center}
\centerline{\includegraphics[width=0.8\textwidth]{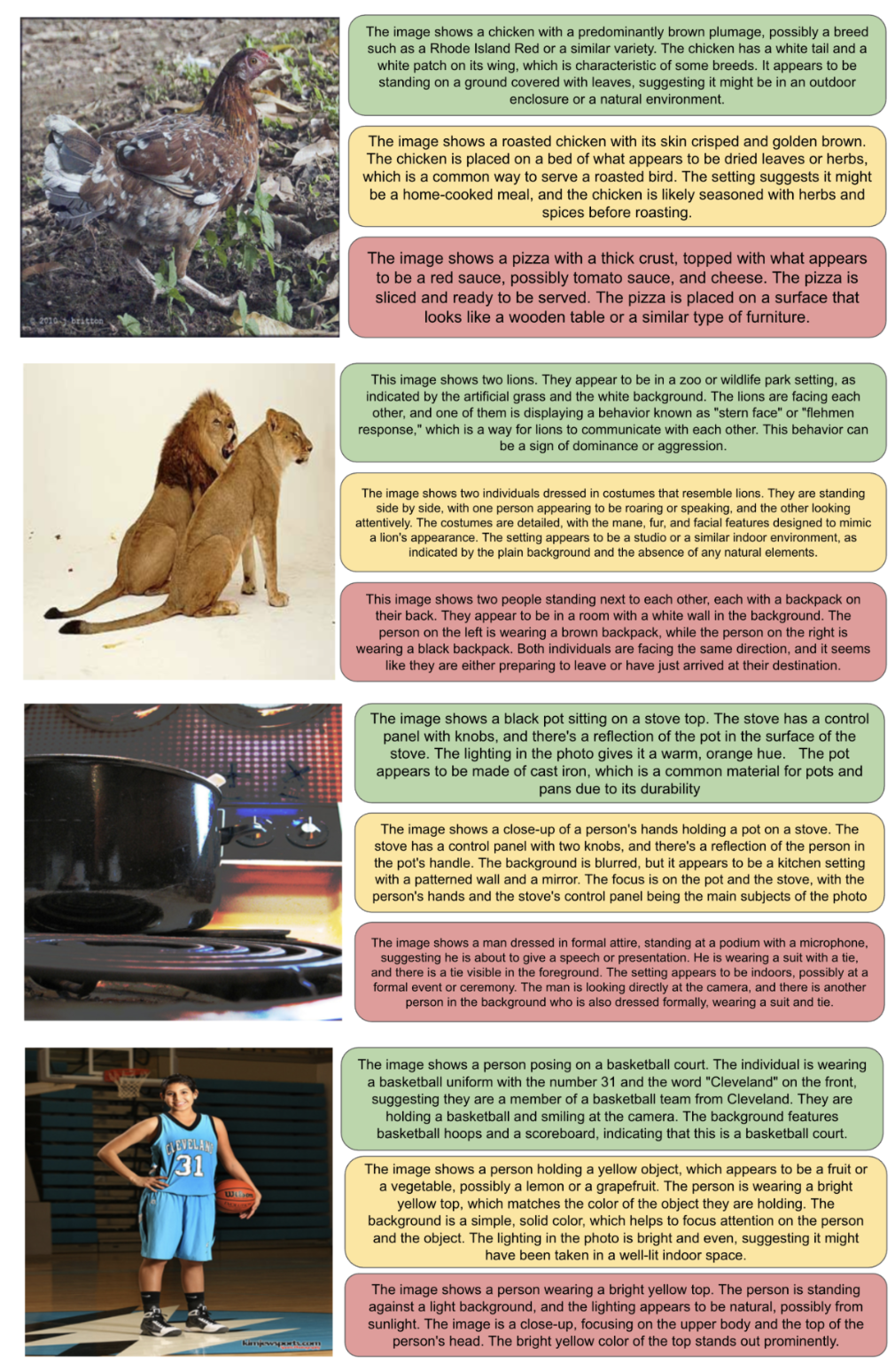}}
\caption{\textbf{Steering Examples} Several selected examples of image-token steering. Three image descriptions are taken, with the simple prompt: ``What is this image?", at three steering levels. The first sample is steered towards ``pizza", the second is steered towards ``backpack", the third is steered towards ``groom", and the fourth is steered towards ``lemon".}
\label{steering_appendix}
\end{center}

\end{figure}

In figure \ref{steering_appendix}, we provide a handful of examples of steered model completions. The steering creates hallucinations of content related to the target class, while maintaining a coherent description of the hallucinated image. Furthermore, at the intermediate steering levels, the original context of the image is maintained: the model hallucinates two humans in lion costumes, instead of two lions, in the second example, which transforms into two people wearing backpacks.

\section{Multimodality Evaluation Details}
\label{mm_eval}
We evaluate our trained SAEs and neurons throughout the model, looking for interpretable and multimodal features and neurons alike. In order to do this, we first sample 25 SAE features and 50 Neurons from layers 8,12,16,20,24, for a total of 250 features. We compute activation deciles for each feature across the ShareGPT dataset, using text and image activations separately. For each feature/neuron, we present the user with the given activation deciles in a ``feature browser." Below is an example browser, showing the image activations of a ``boots" SAE feature.

\begin{figure}[!htp]

\begin{center}
\centerline{\includegraphics[width=\textwidth]{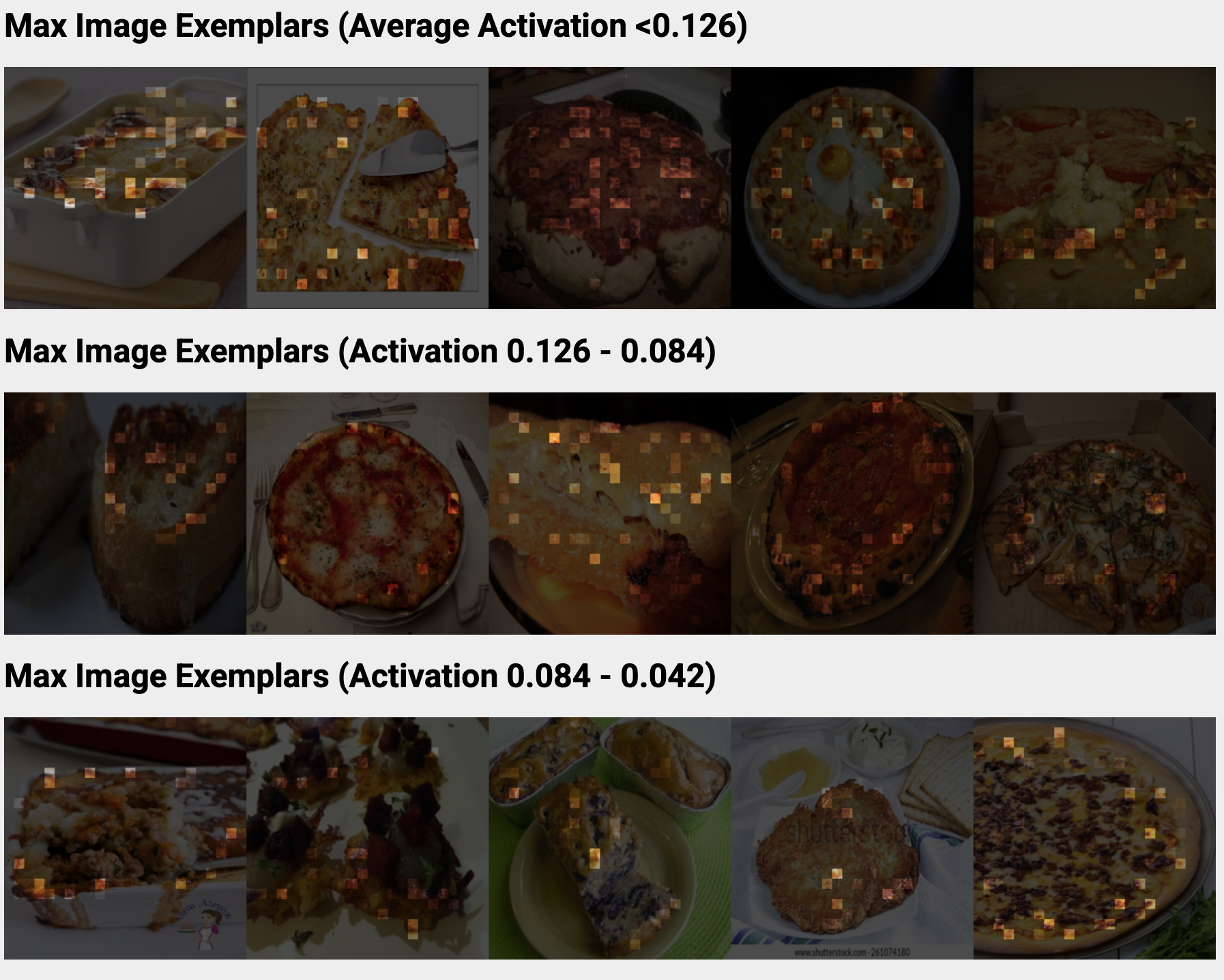}}
\caption{\textbf{Example of a browser:} This SAE feature appears to activate on images of crust and baked goods.}
\label{text_intervention}
\end{center}

\end{figure}

We ask the user to provide text and image interpretations of each feature independently or to respond that the features are uninterpretable. Then, we ask the user if their descriptions of the text and image activation patterns match, to determine feature multimodality.

\section{Adversarial Attacks on VLLMs}

\label{Adversarial Attacks}
\begin{figure}[H]

\includegraphics[width=\textwidth]{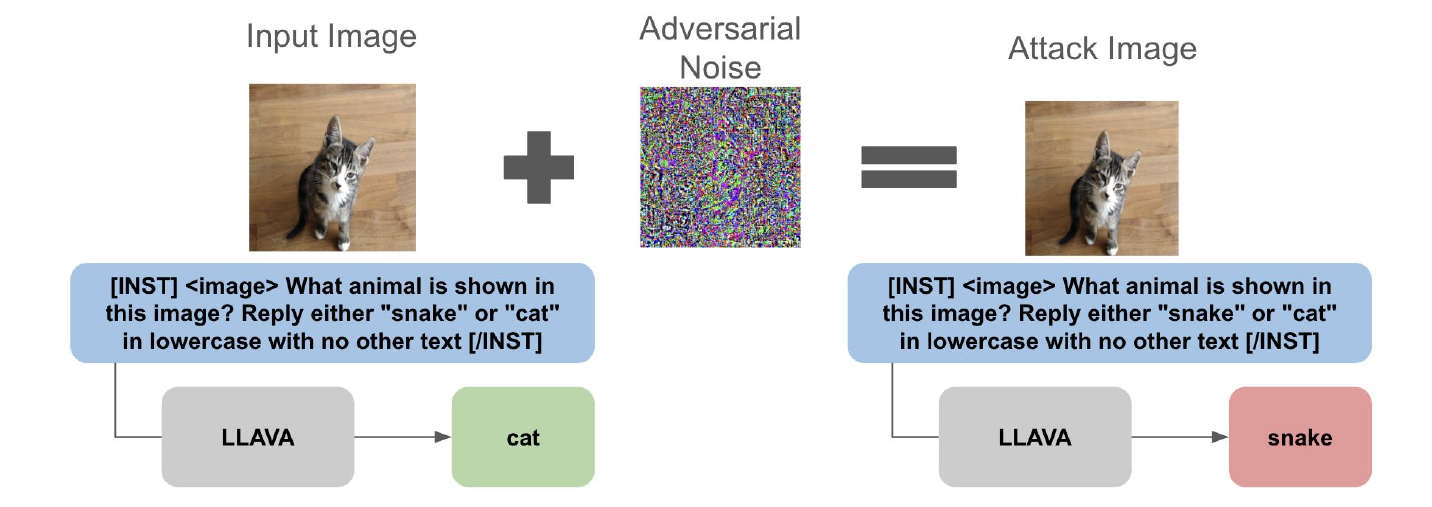}

\end{figure}
Although we find that the image understanding of the model is highly steerable through linear representations, we wonder how other interventions are expressed in the model representations. To do this, we study the setting of Adversarial Attacks, using gradient descent to optimize perturbations to edit the model response. Specifically, we use a single prompt, and optimize the difference in log-probs between an initial and target class using PGD, as shown in section \ref{Adversarial Attacks}.

We ask, how does the Adversarial Attack change model internals? Although we hoped to find that the adversarial attack was well-explained by the SAE basis, we found that the difference in activation between the original and adversarially attacked images were usually \textbf{dense}, involving hundreds of SAE features. Furthermore, the SAE features which changed the most between the images appeared to be random, with them not directly corresponding to the change of the adversarial attack.

Furthermore, we found that adversarial attacks were quite ``brittle", with noise which succeeded in editing model output on one prompt not generalizing well to other prompts. This made experimentation quite tricky. We thus urge further work to consider this setting.

\end{document}